\documentclass[sigconf]{acmart}
\AtBeginDocument{%
  \providecommand\BibTeX{{%
    \normalfont B\kern-0.5em{\scshape i\kern-0.25em b}\kern-0.8em\TeX}}}

\setcopyright{acmlicensed}
\copyrightyear{2024}
\acmYear{2024}
\acmDOI{XXXXXXX.XXXXXXX}

\acmConference[Conference acronym 'XX]{Make sure to enter the correct
  conference title from your rights confirmation emai}{June 03--05,
  2024}{Woodstock, NY}
\acmISBN{978-1-4503-XXXX-X/18/06}

\usepackage{multirow}
\usepackage{enumitem}
\usepackage[bottom]{footmisc}
\setlist{nosep}
\begin{document}
% \title{Multi-Dimensional Hierarchical Relationship Modeling for Click-Through Rate Prediction}
\title{HMDN: Hierarchical Multi-Distribution Network for Click-Through Rate Prediction}

\author{
{Xingyu Lou\textsuperscript{1\textasteriskcentered{}},
Yu Yang\textsuperscript{2\textasteriskcentered{}},
Kuiyao Dong\textsuperscript{1}, Heyuan Huang\textsuperscript{1}, Wenyi Yu\textsuperscript{1}, \\ 
Ping Wang\textsuperscript{1}, Xiu Li\textsuperscript{2}, Jun Wang\textsuperscript{1†}}
}
% \authornote{%
% Both authors contributed equally to this research. Yu Yang performed this work as an intern at OPPO
% }
% \authornote{Corresponding author.}

% \def\thefootnote{*}\footnotetext{These authors contributed equally to this work}
\affiliation{
  \country{{\textsuperscript{1}OPPO, Shenzhen, China}}
}
\affiliation{
  \country{\textsuperscript{2}Shenzhen International Graduate School, Tsinghua University Shenzhen, China}
}
\email{{louxingyu, dongkuiyao, huangheyuan2, yuwenyi, ping.wang}@oppo.com,yu-yang23@mails.singhua.edu.cn,li.xiu@sz.tsinghua.edu.cn,junwang.lu@gmail.com}

\renewcommand{\shortauthors}{Xingyu Lou}

\begin{abstract}
\begin{sloppypar}
As the recommendation service needs to address increasingly diverse distributions, such as multi-population, multi-scenario, multi-target, and multi-interest, more and more recent works have focused on multi-distribution modeling and achieved great progress. However, most of them only consider modeling in a single multi-distribution manner, ignoring that mixed multi-distributions often coexist and form hierarchical relationships. To address these challenges, we propose a flexible modeling paradigm, named \textbf{H}ierarchical \textbf{M}ulti-\textbf{D}istribution \textbf{N}etwork (HMDN), which efficiently models these hierarchical relationships and can seamlessly integrate with existing multi-distribution methods, such as Mixture-of-Experts (MoE) and Dynamic-Weight (DW) models. Specifically, we first design a hierarchical multi-distribution representation refinement module, employing a multi-level residual quantization to obtain fine-grained hierarchical representation.
Then, the refined hierarchical representation is 
% flexibly utilized to guide the model in integrally capturing the hierarchical relationships across multiple multi-distributions.  
integrated into the existing single multi-distribution models, seamlessly expanding them into mixed multi-distribution models.
Experimental results on both public and industrial datasets validate the effectiveness and flexibility of HMDN.
\end{sloppypar}
\end{abstract}


\begin{CCSXML}
<ccs2012>
   <concept>
       <concept_id>10002951.10003317.10003347.10003350</concept_id>
       <concept_desc>Information systems~Recommender systems</concept_desc>
       <concept_significance>500</concept_significance>
       </concept>
 </ccs2012>
\end{CCSXML}

\ccsdesc[500]{Information systems~Recommender systems}


\keywords{Recommender System, Click-Trough Rate Prediction, Hierarchical Representation, Multi-Distribution Modeling}



\maketitle
\section{Introduction}

Online recommendation systems are critical in optimizing user experience and increasing platform revenue by accurately predicting the Click-Through Rate (CTR).
As the business domain expands, it encounters diverse populations, scenarios, and targets, each requiring a tailored recommendation strategy.
However, taking scenarios as an example, building a separate model for each distribution is cost-prohibitive, thus demanding a model that can accommodate multiple scenarios simultaneously. 
All of these can be summarized as a unified modeling paradigm for multiple distributions, which aims to capture the commonalities and differences between distributions.
% Recent research has aimed to develop a unified paradigm that can integrate multiple distributional recommendation needs.

Recently, significant progress has been made in areas such as multi-scenario \cite{star,causalint,samd}, multi-target \cite{mmoe,ple}, multi-population\cite{poso}, and multi-interest\cite{mind,dmin,twin}. 
% For multi-scenario: STAR designed a star-shaped topology to model the commonalities and differences between scenarios through shared and private parameters; CausalInt introduced a causal intervention mechanism to build a scenario-aware estimator within a unified model. For multi-objective: MMOE and PLE use multiple experts to model different data distributions and integrate predictions for different tasks through a gating network; PEPNet constructs a personalized network through a dynamic weighting mechanism. For multi-population: POSO clusters users by activity level, learning different data distributions for different groups. For multi-interest: MIND utilizes dynamic routing and attention mechanisms to guide the learning of users' multi-interest expressions; TWIN captures users' long-term interests through a two-stage consistent lifelong behavior sequence modeling approach.
However, most of these methods only consider modeling in a \textbf{single multi-distribution} manner. In practice, multiple multi-distributions may coexist and form hierarchical relationships, which we refer to as \textbf{mixed multi-distributions}. For instance, in a large e-commerce platform, different populations entering different scenarios often exhibit different interests and intentions. 
% Therefore, it is necessary to carefully design the model for hierarchical modeling various multi-distribution problems.
Therefore, designing carefully to capture the hierarchical relationships within mixed multi-distributions is necessary.
Some recent works have begun to address both multi-scenario and multi-objective issues \cite{pepnet,hinet}. Still, they typically use complex stacking structures, directly stacking the underlying multi-scenario module with the top-level multi-task module. Despite their effectiveness, there are two limitations: 1) when modeling more than two multi-distribution problems simultaneously, simply stacking leads to excessive complexity and thus results in low flexibility; 2) they fail to consider the hierarchical relationships within mixed multi-distributions, which results in sub-optimal performance.

In this paper, we propose a flexible modeling paradigm, named \textbf{H}ierarchical \textbf{M}ulti-\textbf{D}istribution \textbf{N}etwork (HMDN), which efficiently models hierarchical relationships within mixed multi-distributions and can be seamlessly integrated with existing multi-distribution modeling methods.
% with flexibility and adaptability to address these issues via explicitly and implicitly modeling hierarchical relationships between multiple distributions.
% For the first time, we integrate multi-scenario, multi-objective, multi-population, and multi-interest mixed multi-distribution problems into a unified paradigm and point out the hierarchical relationships between different types of multi-distributions. 
By leveraging these hierarchical relationships, we can associate parameters between different data distributions, modeling a more efficient unified structure. Specifically, we first introduce a Hierarchical Multi-Distribution Representation Refining (HMDRR) module, which recursively applies quantization on residuals at multiple levels to obtain quantized embeddings that capture these hierarchical relationships from coarse-to-fine granularity. Then, we sum up the extracted 
% mixed multi-distribution embeddings
quantized embeddings to derive the overall hierarchical representation. This hierarchical representation can subsequently integrate into existing single multi-distribution models like Mixture-of-Experts (MoE) models \cite{moe,mmoe,ple} and Dynamic-Weight models (DW), enabling them to model hierarchical relationships within mixed multi-distributions.
In summary, the main contributions of this paper are as follows:
\begin{itemize}
 % \item We propose for the first time a unified modeling problem of multiple multi-distribution data, integrating multi-scenario, multi-objective, multi-population, and multi-interest multi-distribution types into one model, considering their dependencies and achieving hierarchical multi-distribution modeling.

 % \item We highlight the necessity of modeling hierarchical relationships within mixed multi-distributions. To our knowledge, this is the first attempt to propose a flexible modeling paradigm capable of capturing hierarchical relationships within mixed multi-distributions.
 \item To our knowledge, this is the first attempt to highlight the necessity and propose a flexible modeling paradigm capable of capturing hierarchical relationships within mixed multi-distributions.
 % Our approach can seamlessly integrate with existing single multi-distribution modeling backbones.

 % \item \textcolor{red}{We introduce a residual quantization method that recursively quantifies the mixed multi-distribution features of the data through a multi-layer codebook mapping, efficiently extracting the hierarchical relationships between different multi-distribution types. Directly using mixed multi-distribution features to guide model learning avoids simple stacking structures, achieving a unified and flexible mixed multi-distribution modeling strategy.}

\item We design a hierarchical multi-distribution representation refining module with multi-level residual quantization to extract hierarchical representation. This module can seamlessly integrate with existing single multi-distribution models, efficiently enabling them to model hierarchical relationships within mixed multi-distributions.
 % \item Experiments on both public and industrial datasets demonstrate the effectiveness of HMDN to capture the hierarchical relationships between mixed multi-distribution. Furthermore, through experiments that substitute diverse backbone networks, we validate the adaptability that substitute diverse backbone networks, we validate the adaptability of HMDN.
 \item We conduct experiments on both public and industrial datasets, demonstrating the effectiveness of HMDN in capturing hierarchical relationships within mixed multi-distributions. Additionally, through the experiments of substituting various backbone networks, we further validate the adaptability of HMDN.
\end{itemize}
\vspace{-2mm}
\section{Methodology}

\begin{figure*}[t]
	\centering
    \vspace{-5mm}
	\setlength{\belowcaptionskip}{-0.0cm}
	\setlength{\abovecaptionskip}{-1.0cm}
	\includegraphics[width=0.88\textwidth]{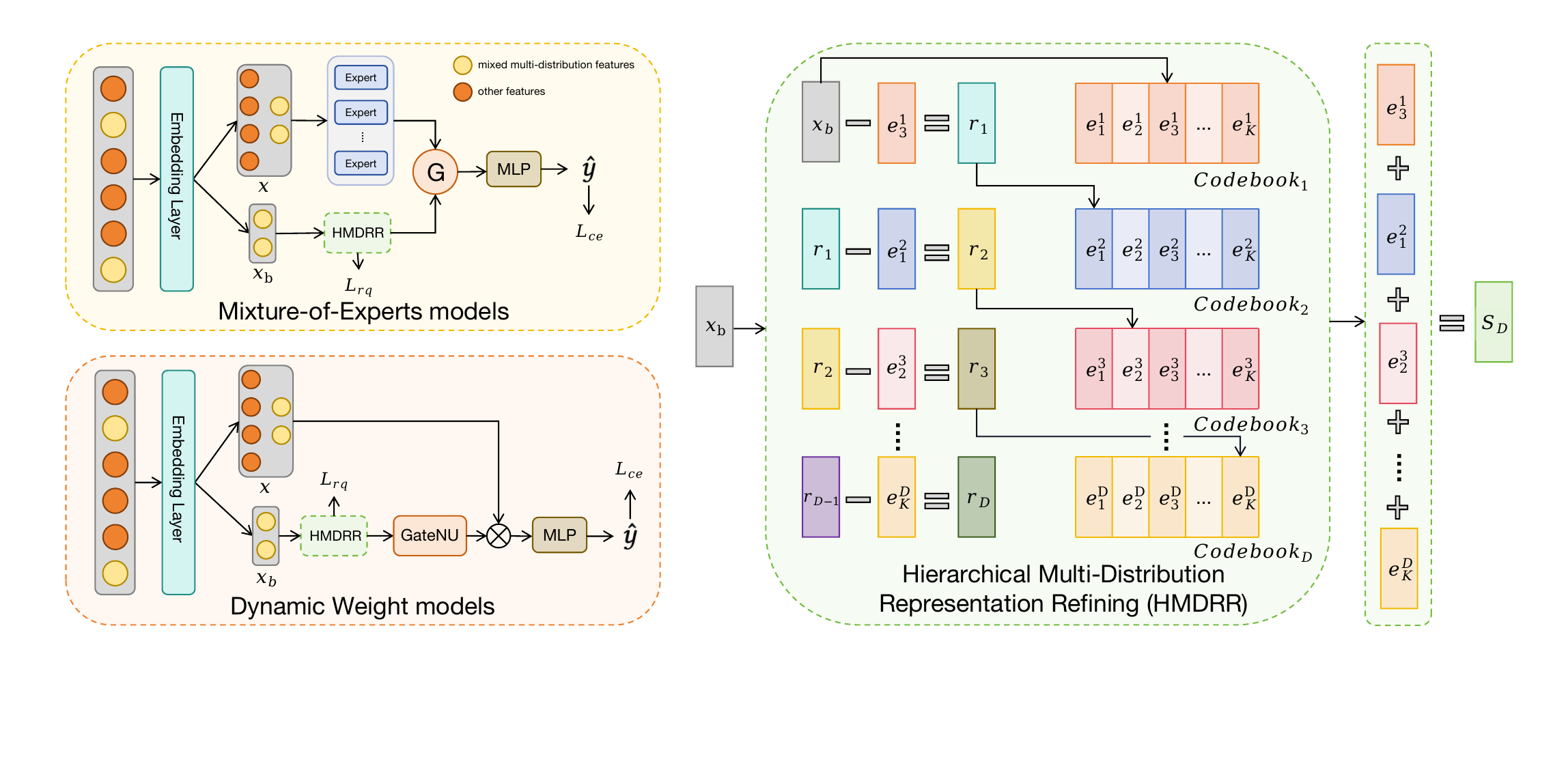}
	\caption{Illustration of the proposed HMDM framework. ($\otimes$) denotes the element-wise multiplication operation. }
	\label{fig:model_frame}
    \vspace{-3mm}
\end{figure*}

\subsection{Problem Formalization}
This paper primarily investigates the unified modeling problem of recommendation systems under mixed multi-distribution data. Given \( N \) different types of multi-distributions \( \{B_1, B_2, \ldots, B_N\} \), for the \( n \)-th multi-distribution, such as multi-scenario, let its distributions be \( \{B_{n1}, B_{n2}, \ldots, B_{nM}\} \), where \( M \) denotes the number of scenarios in the multi-scenario problem. For a set of mixed multi-distribution data \( \{B_{nm}\}, \, m \in [1, M], \, n \in [1, N] \), they share a common feature space $\mathcal{X}$ and a common label space $\mathcal{Y}$ . The goal is to construct a unified CTR prediction function  $f: \mathcal{X} \rightarrow \mathcal{Y}$ that can fully explore the hierarchical relationships within mixed multi-distributions and accurately predict CTR results.
Let the input feature be \( x \), from which we extract the mixed multi-distribution feature \( x_b \) to guide the construction of the unified paradigm.

\subsection{Architecture Overview}
% We propose a unified prediction model for mixed multi-distributions that captures their hierarchical relationships. 
Figure \ref{fig:model_frame} illustrates the complete process of the novel mixed multi-distribution modeling paradigm HMDN.
% As illustrated in Figure \ref{fig:model_frame}, we construct a 
Specifically, for a batch of inputs with various features, we first map the whole input features and mixed multi-distribution features to \( x \) and \( x_b \), respectively, through embedding layers. Then, the Hierarchical Multi-Distribution Representation Refining (HMDRR) module is designed to extract hierarchical representation through multi-level residual quantization.
% constructing layered codebook mappings to quantize the mixed multi-distribution features. 
Finally, the hierarchical representation can flexibly integrate into existing backbone models to guide the model in integrally capturing the hierarchical relationships within mixed multi-distributions. 

\subsection{Hierarchical Multi-Distribution Representation Refining}
\subsubsection{Embedding Layer}
    % In recommendation systems, input data comprises various feature types. For instance, in the Ali-CCP dataset, features include product details, user behavior, scenario specifics, user categories and so on, some linked to multiple distribution traits.

In the embedding layer, we initially map the features relevant to multi-distribution into low-dimensional dense vectors and concatenate them as \( x_b \). Then, we map whole input features to low-dimensional dense vectors \( x \). These representations (\( x \) and \( x_b \)) enable capturing both overall and distribution-specific attributes, thereby enhancing the model's capacity to handle diverse and intricate distributions effectively.

\subsubsection{Residual Quantization Layer}
A multi-level residual quantization is exploited on \( x_b \) to obtain fine-grained hierarchical multi-distribution representation. 
This process allows us to capture nuanced variations and dependencies within the multi-distribution data, resulting in a more precise and detailed representation. Next, we will introduce more details.
 
In vector quantization (VQ) \cite{VQ}, a codebook \( C \) contains codes \( c \) and their embeddings \( e_c \), forming a finite set \( \{(c, e_c)\}_{c \in [K]} \), where \( K \) is the size of the codebook and \( e_c \in \mathbb{R}^{z} \) with \( z \) being the embedding dimension. Given a vector \( s \in \mathbb{R}^{z} \), the vector quantization of \( s \), denoted as \( VQ(s;C) \), finds the code in the codebook whose embedding is closest to \( s \), expressed as:
\begin{equation}    
VQ(s; C) = \text{argmin}_{c \in [K]} ||s - e_c||^2_2.
\end{equation}
However, this may cause the loss of information about hierarchical relationships, as simply merging representations from different levels can overlook nuanced relationships within mixed multi-distributions. 

Therefore, inspired by recent works \cite{Tiger, UIST}, we employ the residual quantization technique, RQ-VAE\cite{RQ}, to capture these hierarchical relationships. 
% partition the single codebook from traditional vector quantization into independent codebooks at each level, connecting them through residual links. 
Let the quantization depth be \( D \). Given the mixed multi-distribution representation \( x_b \), the residual quantization of \( x_b \) can be represented as 
\begin{equation}
RQ(x_b;C,D) = (c_1, ..., c_D),
\end{equation}
where \( c_d \) represents the code at level \( d \). Denoting the residual at level 0 as \( r_0 = x_b \), RQ recursively computes \( c_d \), the code for the residual \( r_{d-1} \), and computes the next residual: 
\begin{equation}
c_d = VQ(r_{d-1};C), \quad r_d = r_{d-1} - e(c_d).
\end{equation}
We sum the quantized embeddings obtained at each level to derive the hierarchical representation vector $s_D$:
\begin{equation}
s_D = \sum_{i=1}^D e(c_i).
\end{equation}
Introducing such a hierarchical structure makes it more suitable for learning mixed multi-distribution representations with dependencies.
% This approach offers several advantages over traditional vector quantization methods. Firstly, by introducing a hierarchical structure, it is more suitable for learning mixed multi-distribution representations with dependencies. 
% \textcolor{red}{Secondly, by using codebook depth instead of increasing codebook size, it helps to conserve space while enhancing the approximation accuracy of quantization.}
It is worth noting that, different from \citet{Tiger} and \citet{UIST}, our goal is to derive the hierarchical representation vector $s_D$, but rather the tuple of codes.

\subsubsection{Dual-View Hierarchical Representation}
% For each depth level \(d\), we construct codebook separately. Compared to codebook sharing, this facilitates extracting hierarchical-specific information, mitigating potential conflicts between levels. 
% Our information extraction methods can be classified into implicit and explicit depending on the input partitioning. 
We design two implementations of representation extraction, implicit and explicit, depending on the input partitioning:

\textbf{Implicit representation extraction.} It directly utilizes the mixed multi-distribution features $x_b$ to establish all codebooks. This is advantageous for autonomously capturing implicit and intricate hierarchical relationships. 

\textbf{Explicit representation extraction.} It entails dividing $x_b$ into $D$ distribution-specific features according to distribution types, for instance, separating features related to populations and scenarios into distinct codebook levels. 
This explicit approach establishes the codebook of each level based on features of one specific distribution type, thereby enhancing interpretability. 
% This explicit approach establishes codebook levels corresponding to distribution types, thereby enhancing interpretability. 

The loss function for the HMDRR module is calculated as follows: 
\begin{equation}
L_{rq} = \sum_{d=1}^D ||e(c_d) - \text{sg}[{x}^{(d)}_b] ||^2_2 + \beta\sum_{d=1}^D ||\text{sg}[e(c_d)] - {x}^{(d)}_b ||^2_2,
\end{equation}
where ${x}^{(d)}_b$represents the multi-distribution embedding input to the \( d \)-th layer codebook, \( \text{sg}[\cdot] \) represents the stop-gradient operator. As \( d \) increases, the quantization error between \( e(c_d) \) and ${x}^{(d)}_b$ gradually decreases, allowing RQ to approximate mixed multi-distribution features from coarse to fine levels while updating the codebook \( C \).

\subsection{Hierarchical Multi-Distribution Modeling}
The proposed Hierarchical Multi-Distribution Representation Refining (HMDRR) module can readily integrate with various backbones. In this section, we implement hierarchical multi-distribution modeling based on MoE-based models and DW-based models.

\textbf{Mixture-of-Experts (MoE).} MoE-based models \cite{moe, mmoe, ple} consists of multiple expert networks and gating networks. The gating networks select the most suitable expert networks for prediction based on the characteristics of the input and combine the outputs of each expert with weights to obtain the final result. The overall formulation is shown as follows:
% \textcolor{red}{It couples data from multiple distributions to generate multiple experts, and then employs gating mechanisms to aggregate experts as the output, making it suitable for handling various distribution scenarios.}
\begin{equation}
\hat{y} = \sum_{i=1}^n G(x)_i E_i(x),
\label{yhatmmoe}
\end{equation}
\begin{equation}
G(x) = \text{softmax}(x \cdot W_g),
\end{equation}
where \( G(\cdot) \) represents the gating network, and \( E_i(\cdot) \) represents the $i$-th expert network.

As discussed above, choosing appropriate input for the gating network is crucial in MoE. An intuitive choice is the multi-distribution representation \( x_b \). However, this is coarse-grained and may lead to the loss of hierarchical distribution information, while isolated distribution representations may exhibit poor generalization to new distributions. Therefore, we use the previously derived hierarchical representation $s_D$ as the gating network input, rather than $x_b$, to facilitate the fine-grained mixed multi-distribution modeling. Equation (\ref{yhatmmoe}) can be reformulated as:

\begin{equation}
\hat{y} = \sum_{i=1}^n G(s_D)_i E_i(x).
\end{equation}

% At this point, \textcolor{red}{MoE-based models can better consider the hierarchical relationships within various distributions, thus enhancing the accuracy and generalization performance.}

\textbf{Dynamic Weight (DW).} DW-based models \cite{pepnet, poso} utilize gating mechanism (e.g. GateNU \cite{pepnet}) to 
% \textcolor{red}{operate on specific representations of distributions, obtaining the gating vector $\delta_b$ under that distribution. }
dynamically scale the bottom-layer embeddings to get the distribution-specific personalized ones:

\begin{equation}
\hat{y} = \delta_b \otimes x,
\end{equation}

\begin{equation}
\delta_b = G_{NU}(x_b),
\end{equation}
where $\otimes$ denotes element-wise multiplication, $G_{NU}(\cdot)$ represent the gating mechanism. 
% Subsequently, an element-wise multiplication between $\delta_b$ and $x$ is introduced to get the distribution-specific representation:

% We introduce the fine-grained Hierarchical Multi-Distribution Representation into the Dynamic Weight Model, replacing the original distribution representations with those quantized through residual quantization as inputs to the GateNU module. 
Similarly, we use the hierarchical representation $s_D$ to replace $x_b$. 
% Thereby, hierarchical multi-distribution information is introduced at the input level.
This allows us to incorporate hierarchical multi-distribution information directly at the input level.

The overall optimization objective is described as follows:
% consists of the RQ loss and the cross-entropy loss:
\begin{equation}
 L =  L_{ce} + {\alpha}L_{rq} = -\frac{1}{N} \sum_{i=1}^{N} \left( y_i \log(\hat{y}_i) + (1 - y_i) \log(1 - \hat{y}_i) \right) + {\alpha}L_{rq} ,
\end{equation}
where the first item is the cross-entropy loss for the CTR task. $\alpha$ denotes the factor to control the importance of $L_{rq}$. \( N \) is the number of samples, $y_i \in \{0,1\}$ is the ground truth of $i$-th sample. \( \hat{y}_i \) is the corresponding predicted probability.

\section{EXPERIMENTS}

\subsection{Experimental Settings}
\subsubsection{Dataset Description}
We conduct extensive experiments on one commonly used public dataset and one industrial dataset, both of which are collected from industrial recommendation systems.

\noindent \textbf{Ali-CCP}. The Ali-CCP dataset is a publicly accessible collection of real-world recommendation system traffic logs in Taobao. It originally contains three domains. We follow the data partitioning of previous work \cite{adaptdhm}, utilizing "\emph{user\_gender}", "\emph{user\_city}" and "\emph{domain\_id}" to partition samples, resulting in $33$ partitions. The training and testing set sizes are over 42.3 million and 43 million, respectively.

\noindent \textbf{Industrial Dataset}. We collect industrial user logs from the OPPO online advertising platform, ranging from 2024-01-02 to 2024-01-09. We divide the training and testing sets chronologically, using the first seven days for training and the last day for testing. We utilize "\emph{domain\_id}", "\emph{is\_new\_user}" and "\emph{ad\_source}" as the partitioner, resulting in $3*2*2 = 12$ partitions. "\emph{is\_new\_user}" and the "\emph{ads\_source}" represent whether it is a new user in a specific domain and the source of the ads (OPPO or third party), respectively.

\subsubsection{Baseline Models}
As mentioned before, our approach is model-agnostic and can integrate with different backbone networks. We select two types of backbone networks for comprehensive comparison. (1) Mixture-of-Experts (MoE) models. These works exploit gate networks to weigh and aggregate multiple expert networks representations according to the input to the gate. Different gate inputs will make the model concentrate on different parts of the experts. We choose MMoE\cite{mmoe}, one of the most representative works among them. (2) Dynamic Weight (DW) models. These works first utilize a gating mechanism to personalize network intermediate outputs through element-wise multiplication between gating vector and hidden units. We choose PEPNet \cite{pepnet}, which has been recently proposed and widely applied. We also compare our approach with the standard baselines used in CTR prediction: DNN\cite{DNN}, DCN\cite{DCN}, Wide\&Deep\cite{WD}, Shared Bottom\cite{SharedBottom}.

\subsubsection{Implementation Details}
For a fair comparison, each MLP in all models has the same hidden layer depth, consisting of 3 layers (128-64-32). The activation function is set to "ReLU". For the hierarchical multi-distribution representation refining process, we tune $\alpha$, the coefficient of $L_{rq}$ in \{0.1,0.5,1.0,2.0\}. We find HMDN is robust to $\alpha$, thus using $\alpha = 1.0$ in all our experiments. Following previous works \cite{Tiger,VQ}, we set $\beta=0.25$. Unless otherwise specified, implicit representation extraction is used by default. For the MoE-based models, we set the number of experts to 3. The optimizer used is Adam \cite{adam} with a learning rate of \( 1 \times 10^{-3} \). 
% All experiments are implemented using Pytorch\footnote{https://pytorch.org/}.
% \subsubsection{Evaluation Metrics}
We use the commonly used metric, AUC (Area under ROC), as the evaluation metric. 
% It is notable that HMDN is a pluggable component that does not introduce any extra hyper-parameters, thus providing higher adaptability and robustness.

\subsection{Experimental Results}
\begin{table}[!t]
    \caption{Performance comparison on Production and Ali-CCP datasets. HMDN (PEPNet) and HMDN (MMoE) denote taking PEPNet and MMoE as backbone, respectively.}
    \centering
    \vspace{-2mm}
    \begin{tabular}{ccccc}
    \toprule
    \multirow{2}{*}{Model} &\multicolumn{2}{c}{Production} &\multicolumn{2}{c}{Ali-CCP} \\
    ~ & AUC & RelaImpr & AUC & RelaImpr \\
    \midrule
    DNN & 0.8206 & +0.00\% & 0.6183 & +0.00\% \\
    DCN & 0.8196 & -0.12\% & 0.6231 & +0.78\% \\
    Wide\&Deep & 0.8328 & +1.48\% & 0.6156 & -0.43\% \\
    \midrule
    Shared Bottom  & 0.8401 & +2.38\% & 0.6188 & +0.08\% \\
    PEPNet & 0.8415 & +2.54\% & 0.6246 & +1.01\% \\
    HMDN (PEPNet) & \textbf{0.8421} &  \textbf{+2.62\%}  & \textbf{0.6250} & \textbf{+1.08\%} \\
    MMoE & 0.8427 & +2.69\% & 0.6237 & +0.87\% \\
    % PEPNet+Exp & 0.7376 & +0.43\% & 0.5866 & +0.32\% \\
    HMDN (MMoE) & \textbf{0.8434} & \textbf{+2.78\%}  & \textbf{0.6253} & \textbf{+1.13\%} \\
    % MMoE+Exp & 0.8428 & +0.43\% & 0.6240 & +0.32\% \\
    \bottomrule
    \end{tabular}
    \label{tab:my_label1}
\end{table}

\begin{table}[!t]
    \caption{The results comparison of different representation extraction types of HMDRR module. "IMP" for implicit and "EXP" for explicit.}
    \centering
    \vspace{-2mm}
    \begin{tabular}{ccccc}
    \toprule
    \multirow{2}{*}{Model} &\multicolumn{2}{c}{Production} &\multicolumn{2}{c}{Ali-CCP} \\
    ~ & AUC & RelaImpr & AUC & RelaImpr \\
    \midrule
    MMoE & 0.8427 & +0.00\% & 0.6237 & +0.00\% \\
    MMoE+EXP & 0.8428 & +0.01\%  & 0.6240 & +0.05\% \\
    % PEPNet+Exp & 0.7376 & +0.43\% & 0.5866 & +0.32\% \\
    MMoE+IMP & \textbf{0.8434} & \textbf{+0.08\%}  & \textbf{0.6253} & \textbf{+0.25\%} \\
    % MMoE+Exp & 0.8428 & +0.43\% & 0.6240 & +0.32\% \\
    \bottomrule
    \end{tabular}
    \label{tab:my_label2}
\end{table}

\subsubsection{Overall Performance}
% 1. effectiveness
% 2. adaptability
Table \ref{tab:my_label1} illustrates the model performances averaged over three times, from which we have several important observations. 
Firstly, single multi-distribution modeling methods such as Shared Bottom, MMoE, and PEPNet outperform traditional methods, indicating that modeling various distributions distinctively is necessary for CTR prediction.
Secondly, comparing MMoE and PEPNet with Shared Bottom reveals that capturing the relationships between distributions at a finer granularity can bring more improvement. Lastly but most importantly, HMDN has achieved the best performance on both public and industrial datasets.
It is notable that the improvements brought by HMDN over MMoE and PEPNet are entirely due to the comprehensive capture of hierarchical relationships within mixed multi-distributions. This only requires minor adaptations to the existing single multi-distribution models, demonstrating the effectiveness and flexibility of HMDN.

\subsubsection{Ablation study}
We further study the impact of different representation extraction types of HMDRR module. Since HMDN with MMoE as its backbone has shown the best performance on both public and industrial datasets, we continue to report results with MMoE as the backbone. Table \ref{tab:my_label2} summarizes the experimental results. Both implicit and explicit representation extraction outperform vanilla MMoE. The improvement in the implicit approach is more significant. The reason could be that implicit representation extraction enables the model for fine-grained modeling, thus uncovering more intricate hierarchical relationships within mixed multi-distributions.

\begin{figure}[!t]
	\centering
    \vspace{-1mm}
	\setlength{\belowcaptionskip}{-0.0cm}
	\setlength{\abovecaptionskip}{-0.0cm}
	\includegraphics[width=0.4\textwidth]{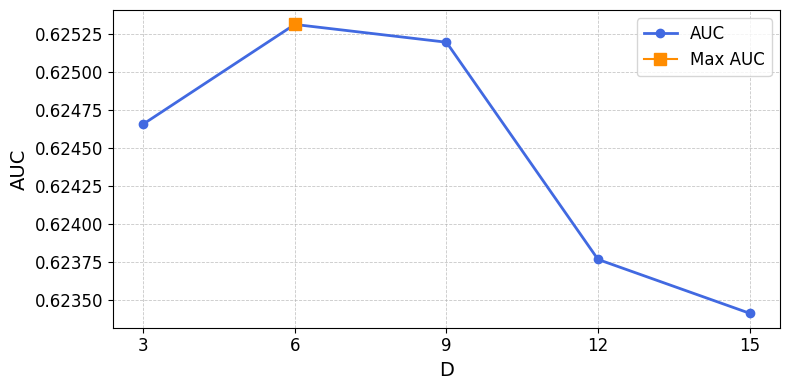}
	\caption{Effect of different codebook depth D.}
	\label{codebook_depth}
    \vspace{-5mm}
\end{figure}

\subsubsection{Hyperparameter Influence}
Due to the limitations of computational resources, we study the impact of codebook depth $D$ on the performance of HMDN under the Ali-CCP dataset, and applied the optimal $D$ to the industrial dataset. From Figure \ref{codebook_depth}, we can observe that HMDN achieves the best performance at $D=6$. 
HMDN exhibits relatively stable performance when the $D$ is between 3 and 9, with differences within 0.06\%, and is superior to the vanilla MMoE (AUC is 0.6237) in all cases. However, when the depth exceeds 9, the performance worsens as the depth increases. This suggests that increasing the depth may not necessarily lead to better results.

\section{CONCLUSION AND FUTURE WORK}
This paper highlights the necessity of modeling hierarchical relationships within mixed multi-distributions and proposes a flexible modeling paradigm named HMDN. HMDN includes a hierarchical multi-distribution representation refining module with multi-level residual quantization to extract hierarchical representation. This module can be seamlessly integrated with existing multi-distribution models, thereby efficiently capturing these hierarchical relationships.
We conduct experiments on both public and industrial datasets to validate HMDN's effectiveness and flexibility. Future research will explore additional applications of the hierarchical representation vector.

\bibliographystyle{ACM-Reference-Format}
\balance
\bibliography{sample-base}

\end{document}